\documentclass[conference]{IEEEtran}
\IEEEoverridecommandlockouts
\usepackage{cite}
\usepackage{amsmath,amssymb,amsfonts}
\usepackage{algorithmic}
\usepackage{url}
\usepackage{multirow} 
\usepackage{makecell}
\usepackage{tabularx}
\usepackage{booktabs}
\usepackage[dvipsnames]{xcolor}
\usepackage{graphicx, pifont}
\usepackage{subfig}
\usepackage{textcomp}
\usepackage{xcolor}
\usepackage{twemojis}

\newcommand{\no}{\scalebox{1}{\twemoji{cross mark}}}
\def\BibTeX{{\rm B\kern-.05em{\sc i\kern-.025em b}\kern-.08em
    T\kern-.1667em\lower.7ex\hbox{E}\kern-.125emX}}
\begin{document}

\title{SpeechLLM Meets Federated Learning for End-to-End ASR: English and Italian Case Studies
}

\author{\IEEEauthorblockN{1\textsuperscript{st} Mohamed Nabih Ali}
\IEEEauthorblockA{\textit{Center of Augmented Intelligence} \\
\textit{Fondazione Bruno Kessler}\\
Trento, Italy \\
mnabih@fbk.eu}
\and
\IEEEauthorblockN{2\textsuperscript{nd} Daniele Falavigna}
\IEEEauthorblockA{\textit{Center of Augmented Intelligence} \\
\textit{Fondazione Bruno Kessler}\\
Trento, Italy \\
falavi@fbk.eu}
\and
\IEEEauthorblockN{3\textsuperscript{rd} Alessio Brutti}
\IEEEauthorblockA{\textit{Center of Augmented Intelligence} \\
\textit{Fondazione Bruno Kessler}\\
Trento, Italy \\
brutti@fbk.eu}

}

\maketitle

\begin{abstract}
Federated learning (FL) enables privacy-preserving training of automatic speech recognition (ASR) systems across distributed data sources, yet its application to large-scale speech language models (SpeechLLMs) remains unexplored. This paper presents the first systematic study of federated training for SpeechLLM-based end-to-end ASR systems. We design a communication-efficient federated optimization strategy tailored to the unique challenges of large speech-language architectures, addressing high-dimensional parameter spaces, gradient communication overhead, and computational constraints in distributed settings. Through extensive empirical evaluation on monolingual ASR tasks in English and Italian, we demonstrate the effectiveness and stability of our federated approach compared to centralized training baselines across diverse acoustic conditions and speaking styles. Additionally, we conduct a comprehensive ablation study analyzing the impact of different speech encoder architectures on monolingual English ASR performance within the federated framework, providing insights into optimal model configurations for decentralized training. Our results achieve competitive word error rates while 
reducing communication costs, establishing practical foundations for federated SpeechLLM deployment in real-world multilingual scenarios.
\end{abstract}


\section{Introduction}
Large Language Models (LLMs) have revolutionized artificial intelligence by demonstrating remarkable capabilities in language understanding, reasoning, and generation across a wide range of applications \cite{hadi2023large}. Their success has recently extended beyond text-only processing to multimodal domains, including speech \cite{zhang2024mm}. In particular, Speech Large Language Models (SpeechLLMs) integrate acoustic modeling with the powerful contextual modeling abilities of LLMs, enabling unified end-to-end systems for automatic speech recognition (ASR), speech translation, and spoken dialogue. Recent frameworks such as SLAM-ASR \cite{ ma2025speech} have demonstrated that incorporating LLM-based architectures into speech recognition pipelines can significantly improve robustness, contextual awareness, and generalization compared to traditional encoder-decoder or CTC-based systems \cite{yoo2025speechllm}. These approaches move toward a unified modeling paradigm where speech and text are processed within a shared representational space, reducing the need for complex cascaded pipelines and task-specific architectures.

Despite these advances, the training of SpeechLLM-based ASR systems typically relies on centralized data collection, where large volumes of speech data are aggregated on remote servers \cite{alhumoud2025asr}. This paradigm raises significant privacy and data governance concerns, particularly in real-world scenarios involving personal devices, call centers, healthcare applications, or multilingual communities \cite{li2024llm}. As an example, speech data is inherently sensitive, often containing biometric information and personal content that may be sensitive for data privacy issues \cite{cheng2024ethical}.

Federated Learning (FL) offers a promising alternative by enabling collaborative model training without transferring raw data to a central server \cite{ali2025efl}. Instead, client devices compute local updates that are aggregated globally, preserving data locality and enhancing privacy \cite{nawar2023fed}. Federated Learning has shown effectiveness in domains such as mobile keyboard prediction and healthcare analytics; however, its application to large-scale SpeechLLM-based ASR remains underexplored \cite{ali2025federating}. Challenges, including client heterogeneity, communication efficiency, and model stability, become even more pronounced when scaling to large speech-language architectures \cite{hamedi2025federated}.

In this work, we propose a federated training framework for end-to-end SpeechLLM-based ASR. Our approach enables decentralized optimization of SpeechLLM models while maintaining competitive recognition performance. We focus on monolingual settings and conduct comprehensive experiments in two languages, English and Italian, to validate the robustness and generality of our method across distinct linguistic conditions. The proposed framework addresses key challenges in federated speech training, including communication constraints and large model adaptation. Through careful optimization strategies and aggregation mechanisms, we demonstrate that SpeechLLMs can be effectively trained in a federated environment without sacrificing accuracy.

The main contributions of this paper are summarized as follows:
\begin{itemize}
    \item We present the first systematic study of federated training for SpeechLLM-based end-to-end ASR.
    \item We propose a communication-efficient federated optimization method for large SpeechLLM models that aggregates only trainable parameters on the server. 
    \item We introduce a modified FedAvg using a unified exponential learning rate decay.
    \item We provide an extensive empirical evaluation on monolingual ASR tasks in English and Italian, demonstrating the effectiveness and stability of the proposed approach.
    \item We analyze the impact of different speech encoders on Monolingual ASR for the English language as an ablation study.
\end{itemize}

\section{Related Work}
In this section, we review the state of the art, first examining FL applied to ASR and related common practices, then discussing the most common communication-efficient approaches for FL.

\subsection{Federated Learning for ASR}
\label{sec:FLASR}
Beyond the standard hurdles of non-IID and imbalanced data, Federated Learning for ASR is further complicated by the intensive resource requirements of contemporary architectures. Models such as Transformers~\cite{zeyer2019comparison}, Transducers~\cite{moriya2023improving, zeineldeen2022conformer}, and RNNs~\cite{oruh2022long} often impose a computational burden that exceeds the hardware constraints of edge devices. Furthermore, ASR traditionally relies on the massive, centralized datasets~\cite{wang-etal-2021-voxpopuli} typically unavailable at the client level. The resulting scarcity of ground-truth labels on client devices necessitates the adoption of unsupervised or self-supervised paradigms.

The literature surrounding FL-based ASR has expanded significantly in recent years. While~\cite{azam2023importance, yu2021federated} offer exhaustive surveys of optimization and training strategies, we focus here on the most pertinent methodological advancements.

Foundational work by~\cite{dimitriadis2020federated} introduced dynamic gradient aggregation specifically for ASR tasks. Subsequent studies, such as~\cite{gao2022federated}, have demonstrated that weighting client updates by Word Error Rate (WER) yields superior performance compared to traditional loss-based weighting. This research also highlights the necessity of a centralized pre-training phase to ensure convergence, supplemented by a post-aggregation training stage to mitigate model divergence across communication rounds findings corroborated by~\cite{gao2022end}. Conversely,~\cite{nguyen2023federated} suggests that effective cross-domain ASR in an FL environment is primarily achievable only through large-scale pre-trained models, such as Wav2Vec2.0~\cite{baevski2020wav2vec}.

To counter the lack of local labels, several works~\cite{jia2022federated, dimitriadis2020federated} have explored self-supervised and unsupervised FL frameworks. However, as the present study concentrates on specific neural architectures for Efficient Federated Learning (EFL), we treat these supervision strategies as orthogonal and applicable to our proposed structural improvements.

\subsection{Communication Efficient FL}
A wide array of techniques has been developed to optimize bandwidth efficiency in distributed systems by minimizing the data volume transmitted to central servers. Sparsification methods, which restrict communication to a strategic subset of model updates—such as top-$K$ gradients~\cite{sattler2019robust,li2021talk,stich2018sparsified}—have been successfully adapted for ASR frameworks~\cite{jia2022federated}. Alternatively, Quantization strategies compress local gradients prior to transmission to reduce bit-rate requirements, though this often necessitates a trade-off between communication gains and potential accuracy loss or increased local computation~\cite{tonellotto2021neural,Yu-FedLora-2023,liu2022hierarchical}. Furthermore, Knowledge Distillation has emerged as a viable solution for reducing exchange overhead by training compact student models to emulate the behavior of high-capacity teacher architectures~\cite{zhu2021data}.

Beyond compression, recent research has shifted toward modifying model parameterization to alleviate resource bottlenecks. Low-rank factorization facilitates the distribution of compact representations to heterogeneous devices for subsequent full-rank global aggregation, a technique explored for CNNs in~\cite{yao2021fedhm}. Similarly, Early-Exit architectures allow clients to transmit only specific model components during the aggregation process~\cite{ali2024federating,lee2024recurrent}.

Recently, Parameter-Efficient Fine-Tuning (PEFT) methodologies including Low-Rank Adaptation (LoRA)~\cite{hu2021lora,zhang2023adaptive} and various adapter modules~\cite{pfeiffer2020adapterfusion,houlsby2019parameter} have proven highly effective for adapting massive pre-trained models like Llama~\cite{touvron2023llama}, Gemini~\cite{team2023gemini}, and Mistral~\cite{jiang2023mistral}. Given that PEFT typically introduces only $1\%$--$2\%$ additional parameters relative to the base model~\cite{babakniya2023slora}, it is exceptionally well-suited for environments with stringent computational and bandwidth constraints. However, it is important to note that PEFT methods can exhibit higher sensitivity to non-IID data distributions than full fine-tuning, necessitating the use of specialized mitigation strategies~\cite{babakniya2023slora}.

Despite advances in communication efficiency, current FL methods are largely optimized for smaller, specialized architectures. Thus, it may struggle to scale to SpeechLLMs. Existing strategies often face convergence instability when applied to billion-parameter architectures on non-IID speech data. This leads to a "client drift," where the global model fails to reconcile the diverse acoustic and linguistic profiles across decentralized devices, resulting in poor accuracy and training volatility.

This work presents the first systematic study of FL for SpeechLLM-based ASR. We propose a communication-efficient framework that aggregates only trainable parameters, significantly reducing bandwidth for large-scale models. By introducing a modified FedAvg with unified exponential learning rate decay, we ensure stable convergence across heterogeneous datasets. Our approach is validated on English and Italian ASR tasks, featuring an ablation study on the impact of various speech encoders.

\section{Federated SpeechLLM Architecture}

\begin{figure*}%
\centering
\subfloat[\centering]{{\includegraphics[width=0.4\linewidth]{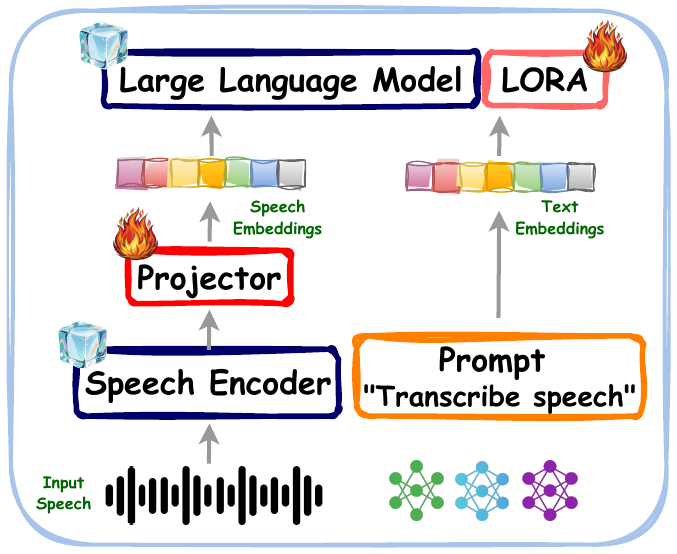} }}%
\quad
\subfloat[\centering]{{\includegraphics[width=0.4\linewidth]{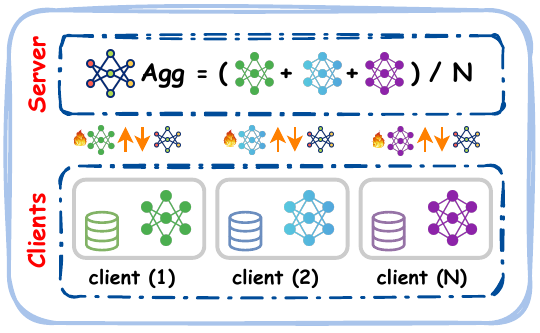} }}%
\caption{(a) Conventional Speech-LLM architecture. (b) Proposed federated Speech-LLM training pipeline.}%
\label{fig:architecture}%
\end{figure*}

The proposed pipeline, illustrated in Figure~\ref{fig:architecture}, consists of two main components: (a) a conventional SpeechLLM-based ASR architecture and (b) the federated training pipeline. In the following sections, we discuss in detail each part of the proposed pipeline.

\subsection{SpeechLLM-Based ASR Architecture}
As shown in Figure~\ref{fig:architecture}(a), the SpeechLLM architecture integrates three components: a speech encoder, a projection layer, and an LLM backbone integrated with LoRA. The speech encoder converts raw waveforms into frame-level acoustic embeddings. The projection layer then maps these embeddings into the LLM's token embedding space. Finally, the LLM performs auto-regressive decoding to generate transcriptions. For efficient adaptation, LoRA modules are integrated into the LLM layers, allowing only the LoRA and projector parameters to be trained while keeping the LLM backbone frozen. In the following sections, we briefly explain each component of the SpeechLLM pipeline.

\subsection{Speech Encoder}
In our experiments we investigate the performance of two common speech encoders \ding{182} WavLM-large and \ding{183} Whisper-medium.

\ding{182} WavLM-Large \cite{chen2022wavlm} is a self-supervised speech encoder developed by Microsoft with 317 million parameters. It uses masked prediction and denoising objectives trained on unlabeled speech data (LibriLight, GigaSpeech, VoxPopuli). The model excels at extracting robust speech representations that capture both content and speaker characteristics, making it effective for downstream tasks like speaker verification, speech separation, and ASR \cite{hu2024wavllm}. Its transformer-based architecture learns generalizable features without requiring labeled data.

\ding{183} Whisper-Medium \cite{radford2023robust} is OpenAI's 769-million parameter multilingual speech recognition model trained on 680,000 hours of weakly-supervised data. It uses an encoder-decoder transformer architecture where the encoder processes mel-spectrograms and the decoder generates text transcriptions. Whisper-Medium supports recognition in 99 languages and translation to English. It demonstrates strong robustness to diverse accents, background noise, and audio quality variations, making it practical for real-world ASR applications.

\subsection{Large Language Model}
In our pipeline, we rely on TinyLlama-1.1B-Chat-v1.0 \cite{zhang2024tinyllama}, a compact open-source language model with 1.1 billion parameters, trained on approximately 3 trillion tokens following the Llama 2 architecture. Despite its small size, it's specifically fine-tuned for conversational interactions using instruction-following and chat datasets. The model is designed for efficient deployment on resource-constrained devices while maintaining reasonable performance on dialogue tasks, question answering, and text generation. TinyLlama uses the same tokenizer and architecture principles as larger Llama models but offers significantly faster inference and lower memory requirements, making it practical for edge computing, mobile applications, and scenarios where computational resources are limited \cite{wang2024efficient}. Contemporary LLMs are frequently fine-tuned using LoRA \cite{zhang2024scaling}. LoRA ~\cite{hu2022lora} introduces trainable low-rank matrices into transformer layers to approximate weight updates. For a pre-trained weight matrix $\mathbf{W} \in \mathbb{R}^{d \times d_k}$, LoRA represents its update with a low-rank decomposition $\mathbf{W} + \Delta \mathbf{W} = \mathbf{W} + \mathbf{AB}$, where $\mathbf{A} \in \mathbb{R}^{d \times r}$, $\mathbf{B} \in \mathbb{R}^{r \times d}$ are learnable and $r \ll d$ \cite{pletenev2025much}. 

\subsection{Projector}
To align the dimensionality between the speech encoder and language model, we use a simple two-stage linear adapter network: a projection layer followed by average pooling \cite{ma2026slam}. The projection layer transforms the speech encoder embeddings into the 2048-dimensional input space of TinyLlama. Subsequently, we apply 1D average pooling with a kernel size and stride of $k=2$ along the time axis, which compresses the sequence length by half. This pooling step reduces computational costs during auto-regressive decoding while consolidating local temporal features into more compact representations. This architecture maintains a minimal parameter footprint, concentrating trainable parameters in the linear transformation while the pooling operation provides parameter-free compression. During training, only the adapter parameters are updated while both the speech encoder and language model remain frozen, allowing efficient bridging between the two pretrained modules.

\subsection{Federated Training Framework}
Figure~\ref{fig:architecture}(b) shows the proposed federated learning framework for decentralized SpeechLLM training. Instead of aggregating speech data on a central server, training is performed collaboratively across multiple clients, each holding local speech datasets.

At each communication round, the central server broadcasts the current global model parameters to a subset of participating clients. Each client performs local training using its private speech data, updating the trainable components of the SpeechLLM (i.e., LoRA and projector layers). The updated local model parameters are then transmitted back to the server. The server aggregates these parameters using a weighted averaging strategy \cite{mcmahan2017communication}:


\begin{equation}
\boldsymbol{\theta}^{(t+1)} =
\sum_{k=1}^{N}
\frac{n_k}{\displaystyle\sum_{i=1}^{N} n_i}
\boldsymbol{\theta}_k^{(t)}
\end{equation}

where $\boldsymbol{\theta}^{(t+1)}$ denotes the updated global model parameters at communication round $t+1$,  $\boldsymbol{\theta}_k^{(t)}$ represents the locally updated model parameters obtained by client $k$ at round $t$, $n_k$ is the number of local training samples at client $k$, $N$ is the total number of participating clients in round $t$, and $\sum_{i=1}^{N} n_i$ represents the total number of training samples across all participating clients. The aggregated model is redistributed in the next communication round.

By restricting parameter updates to lightweight adaptation modules, the proposed framework reduces communication costs and improves scalability when training large SpeechLLMs in federated environments. Furthermore, since raw speech data never leaves client devices, the framework enhances privacy preservation while enabling collaborative learning across geographically distributed users.

Overall, the proposed architecture bridges SpeechLLM modeling and federated optimization, enabling privacy-aware, decentralized training for end-to-end ASR systems.

\section{Experimental Setup}
\subsection{Dataset}
For English, we utilize the LibriSpeech-100 dataset \textbf{(LS)} \cite{panayotov2015librispeech}, consisting of approximately 100 hours of read English speech from LibriVox audiobooks with high-quality, time-aligned transcriptions. We use the \textit{train-clean-100} split for training and the standard \textit{test-clean} set for evaluation. For Italian, we use the Italian portion of the Multilingual LibriSpeech \textbf{(MLS)} corpus \cite{pratap2020mls}, comprising several hundred hours of read Italian speech with corresponding transcriptions from public-domain LibriVox audiobooks. This dataset enables evaluation of multilingual ASR models in a non-English scenario. Table \ref{tab:stat} summarizes the statistics of both datasets in terms of hours and speakers.

\subsection{FedAvg with learning rate scheduler (Adaptive FedAvg)}
The proposed federated learning strategy introduces key enhancements over vanilla FedAvg \cite{mcmahan2017communication}, making it well-suited for large-scale speech model adaptation. Rather than assigning static and heterogeneous learning rates across client subsets, the modified strategy adopts a unified exponential learning rate decay schedule defined as:
\begin{equation}
    \eta_t = \eta_0 \cdot \gamma^{\lfloor t / \tau \rfloor}
\end{equation}
where $\eta_0 = 0.001$ is the initial learning rate, $\gamma = 0.9$ is the decay factor, $\tau = 10$ is the decay period in rounds, and $t$ is the current federated round. This promotes training stability and smoother convergence, as clients collectively transition from aggressive early-round updates to more refined late-round refinements.

\subsection{Federated Learning Setup}
We implement our FL framework using Flower \cite{beutel2020flower}. We deploy a client for each speaker in the datasets. In each round, $30\%$ of the clients are randomly instantiated for 10 epochs of local training. The resulting gradients are centrally agglomerated using \textbf{FedAvg} strategy \cite{mcmahan2017communication}. FL is applied for 100 rounds. All FL results marked as \textcolor{orange}{\textbf{—-}} , \textcolor{ForestGreen}{\textbf{—-}} for LS and MLS, respectively, are compared with those achieved with central training marked as \textcolor{blue}{$\star$} , \textcolor{red}{$\star$} for LS and MLS, respectively. More details related to training hyperparameters are publicly available in our GitHub repository \footnote{
\url{https://github.com/mnabihali/Fed-SpeechLLM}}.

\begin{table}
\centering
\caption{Statistics (duration in hours, and the number of speakers) of LibriSpeech and MLS datasets.}
\label{tab:stat}
\begin{tabular}{lccccc}
\toprule
 & \multicolumn{2}{c}{LibriSpeech} & \multicolumn{2}{c}{MLS} \\
\cmidrule(lr){2-3} \cmidrule(lr){4-5}
 & hours & \# Spks. & hours & \# Spks. \\
\midrule
Train & 100 & 251 & 247.38 & 65 \\
Test  & 5.4 & 40 & 5.27 & 10 \\
\bottomrule
\end{tabular}
\end{table}

\section{Experimental Results}
\subsection{Adaptive FedAvg}
We conduct a comparative evaluation of both vanilla and adaptive FedAvg on the LS dataset, employing WavLM as the speech encoder. Model performance is quantified using WER to assess recognition accuracy rigorously.

\begin{figure}
    \centering
    \includegraphics[width=\linewidth]{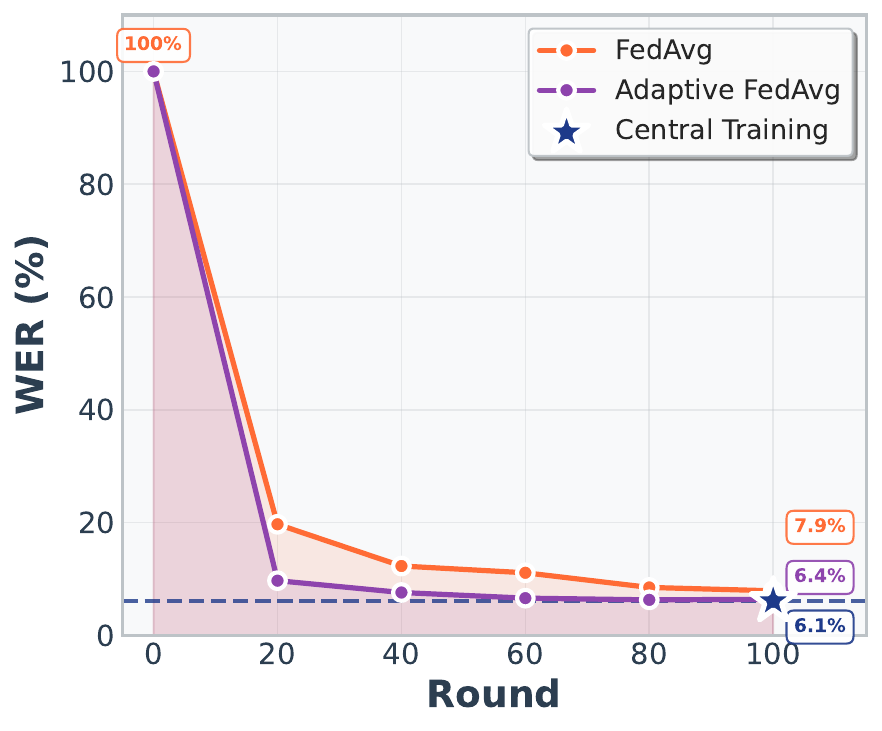}
    \caption{WER comparison between FedAvg and Adaptive FedAvg on the LS dataset using WavLM as speech encoder. The dashed line indicates the central training reference (6.1\%).}
    \label{fig:adaptfedavg}
\end{figure}

The WER reported in Figure \ref{fig:adaptfedavg} across 100 federated rounds confirm the superiority of the proposed Adaptive FedAvg strategy. At round 20, Adaptive FedAvg 
achieves a WER of 9.7\% compared to 19.7\% for standard FedAvg with a relative improvement of nearly 51\%. This demonstrates significantly faster early convergence. By round 100, the proposed strategy attains a final WER of 6.4\% versus 7.9\% for the baseline, approaching the central training reference of 6.1\%. These results confirm that the exponential learning rate decay yields both faster convergence and a stronger final model,  underscoring the practical value of Adaptive FedAvg for federated speech model adaptation.

\subsection{Monolingual FL with WavLM}
\begin{figure*}
    \centering
    \subfloat[\centering]{{\includegraphics[width=0.45\linewidth]{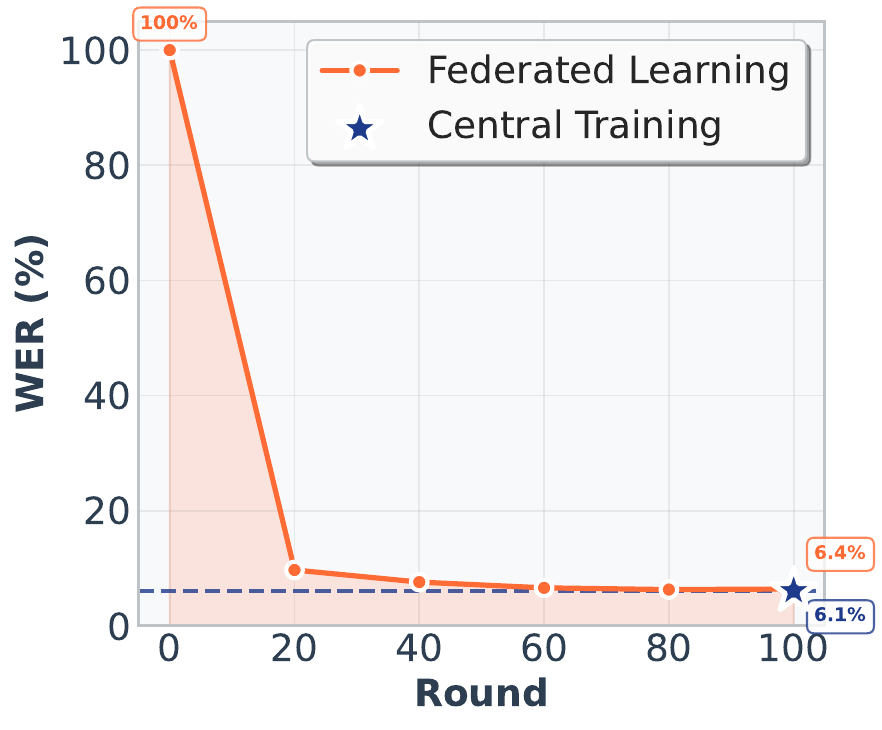} }}%
    \quad
    \subfloat[\centering]{{\includegraphics[width=0.45\linewidth]{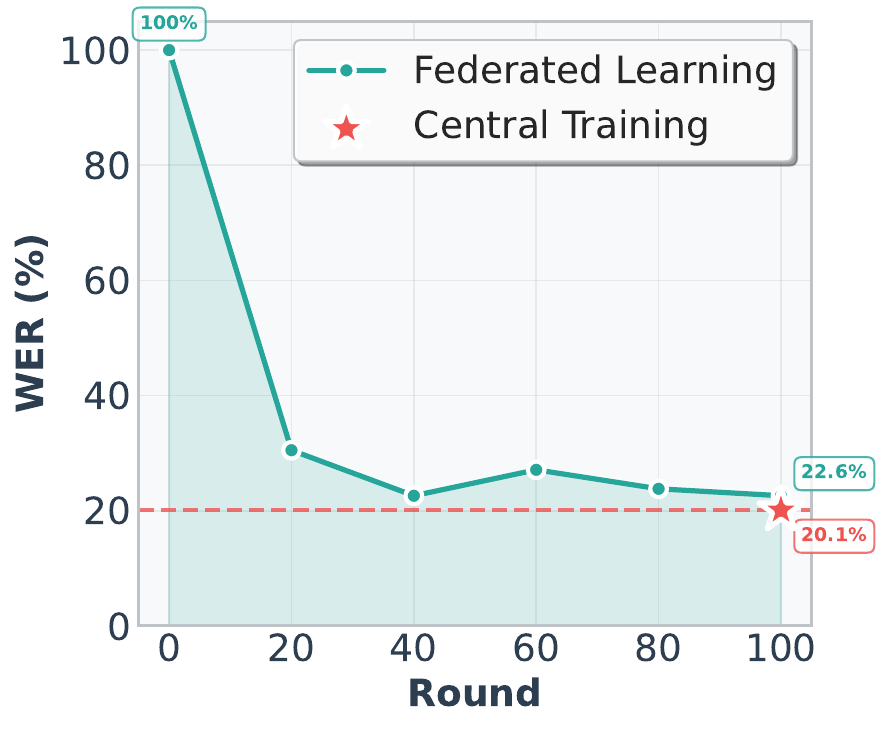} }}%
    \caption{WER in Monolingual scenario with SpeechLLM in both central and Federated scenarios using WavLM as speech encoder. (a) LS English dataset. (b) MLS Italian dataset.}%
    \label{fig:WavLM}%
\end{figure*}

In this scenario, we utilize WavLM as a speech encoder for the SpeechLLM architecture for both federated and centralized approaches on  LS and MLS monolingual datasets, evaluating convergence behavior over 100 federated rounds. Figure \ref{fig:WavLM} presents the WER trajectories, revealing distinct convergence patterns and language-specific challenges in federated speech recognition systems.

In the case of the LS dataset, as depicted in Figure \ref{fig:WavLM}(a), FL demonstrates remarkable convergence characteristics, starting from an initial WER of $\approx$ 100\% and rapidly improving to approximately 10\% by round 20. The steepest performance gains occur within the first 20-40 rounds, after which the learning curve stabilizes. By round 100, the FL pipeline achieves a WER of 6.4\%, effectively matching the centralized training baseline of approximately 6.1\%. This convergence demonstrates that federated optimization can successfully aggregate gradients across distributed clients to reach performance parity with centralized training, despite the challenges of non-collocated data and communication constraints.

Regarding the Italian MLS dataset, as shown in Figure \ref{fig:WavLM}(b), exhibits more challenging federated learning dynamics. While the convergence pattern follows a similar trajectory starting at 100\% WER and rapidly decreasing through early rounds, a persistent performance gap remains throughout training. The federated model achieves approximately 22\% WER by round 100, compared to 20.1\% WER for centralized training. This absolute WER difference $\approx$ 2\% highlights the robustness and strong generalization capability of the federated approach, even under more challenging data conditions. The language-dependent performance gap between English and Italian federated training highlights the impact of data characteristics on federated optimization effectiveness. 

From a communication efficiency perspective, both experiments demonstrate that meaningful convergence occurs within 60-80 federated rounds, with diminishing returns beyond this point. This finding is practically significant, as it establishes a reasonable communication budget for federated SpeechLLM training. 

\subsection{Monolingual FL  with Whisper}

\begin{figure*}
 \centering
    \subfloat[\centering]{{\includegraphics[width=0.45\linewidth]{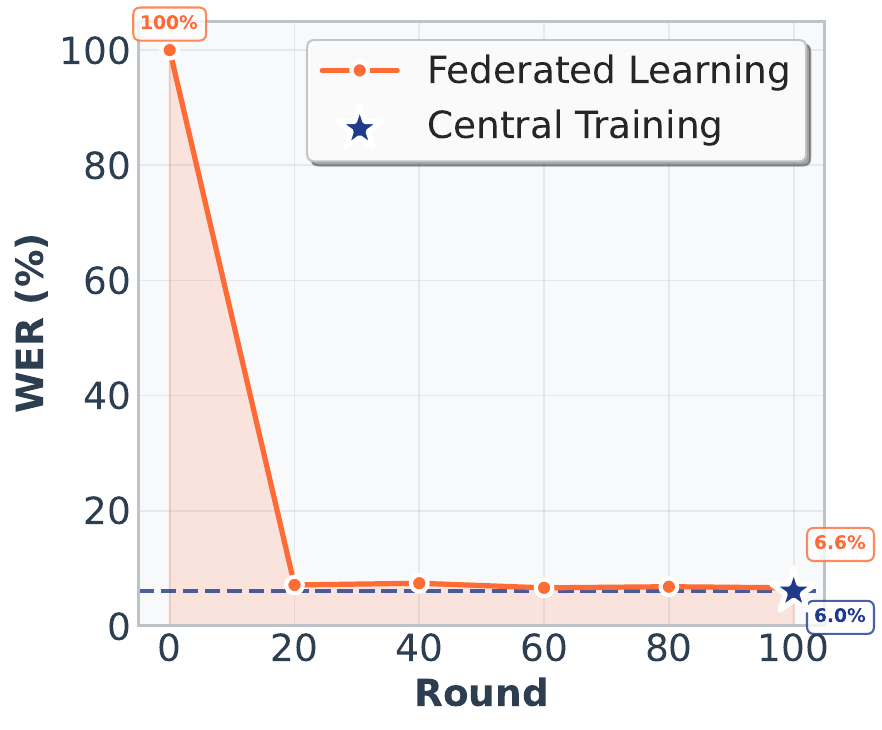} }}%
    \quad
    \subfloat[\centering]{{\includegraphics[width=0.45\linewidth]{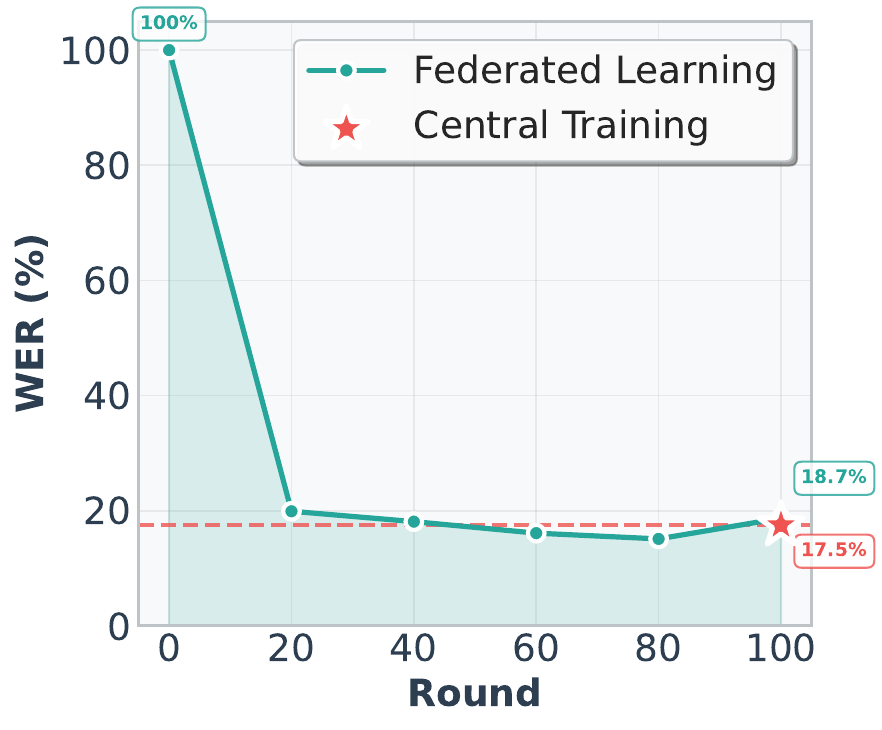} }}%
    \caption{WER in Monolingual scenario with SpeechLLM in both central and Federated scenarios using Whisper as speech encoder. (a) LS English dataset. (b) MLS Italian dataset.}%
    \label{fig:whisper}%
    
\end{figure*}
In this scenario, we investigate the Whisper model as a speech encoder, evaluated under both federated and centralized optimization strategies, with the same hyperparameters and datasets as the previous experiment. Figure~\ref{fig:whisper} illustrates the WER trends for both monolingual datasets.

For LS dataset (Figure~\ref{fig:whisper}(a), FL shows rapid convergence, reducing WER from 100\% to approximately 7\% by round 20. The performance then stabilizes, reaching 6.6\% at round 100, closely matching the performance of the centralized baseline of 6.0\%. The small absolute gap of 0.6\% demonstrates that federated optimization can effectively preserve Whisper’s pretrained representations and achieve near-centralized performance.

For the MLS Italian dataset (Figure~\ref{fig:whisper}(b), convergence follows a similar trend but with a slightly larger gap. The federated model decreases from 100\% WER to $\approx$ 20\% within the first 20 rounds and continues improving gradually. By round 100, federated training achieves 18.7\% WER compared to 17.5\% under centralized training, resulting in a modest 1.2\% absolute difference. 

Overall, the results indicate that Whisper enables stable and efficient federated fine-tuning, substantially reducing the centralized--federated performance gap. Moreover, most performance gains occur within the first 40--60 rounds, suggesting that effective convergence can be achieved with a limited communication budget.

Compared to WavLM, the Whisper-based federated model exhibits greater robustness to client heterogeneity and sustains a consistently smaller performance gap relative to centralized training, particularly on the MLS dataset. This suggests that Whisper’s pretrained representations and stronger cross-lingual generalization capabilities better accommodate the statistical variability inherent in MLS, leading to more stable convergence and enhanced performance under decentralized optimization settings.

\subsection{SpeechLLM versus PEFT}
Table 2 highlights the impact of model parameterization on optimization stability and performance under centralized and FL using the LS dataset. While full fine-tuning of WavLM (85.1M parameters) achieves the best centralized WER (4.4\%), this approach requires updating all model parameters and is therefore ill-suited for federated settings due to excessive communication cost and optimization instability. In contrast, parameter-efficient methods reduce the number of trainable parameters by over 89\% (EL-adapters: 9.1M) and 90\% (Speech-LLM: 8.4M) relative to full fine-tuning, while maintaining competitive recognition accuracy.

In the federated regime, full WavLM fine-tuning fails to converge, confirming the difficulty of optimizing large acoustic models under non-IID data distributions and limited client communication. Adapter-based approaches remain viable, with federated adapters achieving a WER of 6.1\%. Notably, Speech-LLM exhibits stable convergence and attains a comparable WER of 6.4\% while updating 90.1\% fewer parameters than WavLM-FT. This demonstrates that the modular speech–language architecture of Speech-LLM is well aligned with federated optimization, enabling effective parameter aggregation without sacrificing robustness.

Overall, these results suggest that LLM-augmented speech models, when combined with parameter-efficient training strategies, can significantly mitigate the communication and optimization challenges of federated learning. Speech-LLM thus offers a practical trade-off between model expressiveness and federated efficiency, narrowing the performance gap between centralized and decentralized ASR while reducing both training and communication overhead.

\begin{table}
\scriptsize
\centering
\caption{Comparison of centralized and federated training setups using PEFT and SpeechLLM approaches.}
\label{tab:speechllm}
\begin{tabular}{cccc}
\toprule
\textbf{Training} & \textbf{Model} & \textbf{\# Params} & \textbf{WER (\%)} \\
\midrule
\multirow{3}{*}{\makecell{Centralized\\Training}}
 & WavLM-FT              & 85.1 M & 4.4 \\
 & WavLM EL-adapters \cite{ali2025efl}       & 9.1 M  & 4.6 \\
 & Speech-LLM
                         & 8.4 M
                                 & 6.1 \\
\midrule
\multirow{3}{*}{\makecell{Federated\\Learning}}
 & WavLM                 & 85.1 M & \no \\
 & WavLM EL-adapters \cite{ali2025efl}       & 9.1 M  & 6.1 \\
 & Speech-LLM
                         & 8.4 M
                                 & 6.4 \\
\bottomrule
\end{tabular}
\end{table}

\subsection{From Monolingual to Multilingual}

\begin{figure*}
    \centering
    \subfloat[\centering]{{\includegraphics[width=0.48\linewidth]{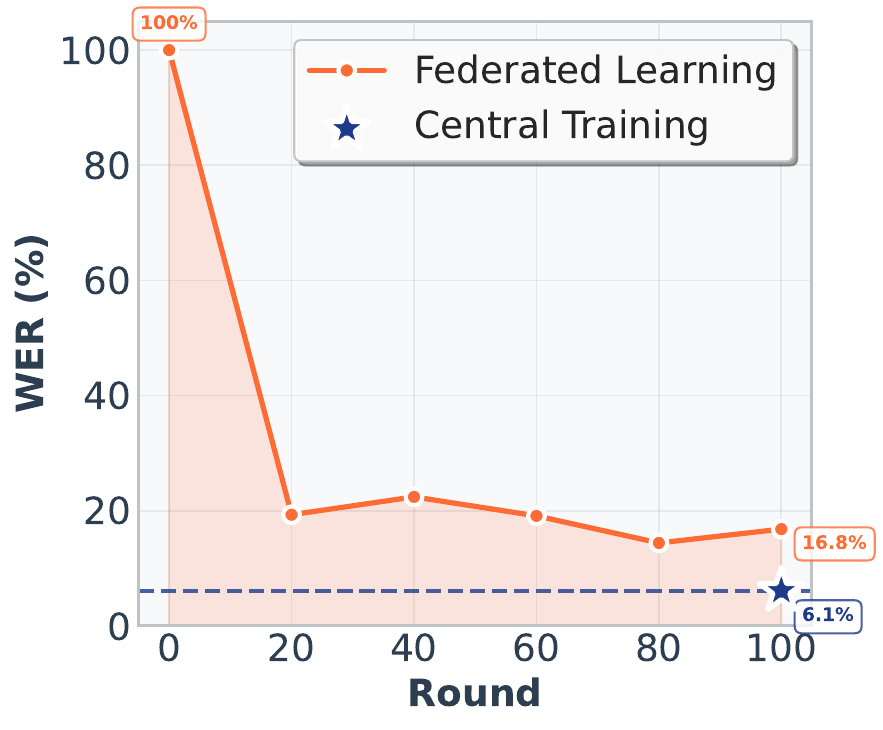} }}%
    \quad
    \subfloat[\centering]{{\includegraphics[width=0.48\linewidth]{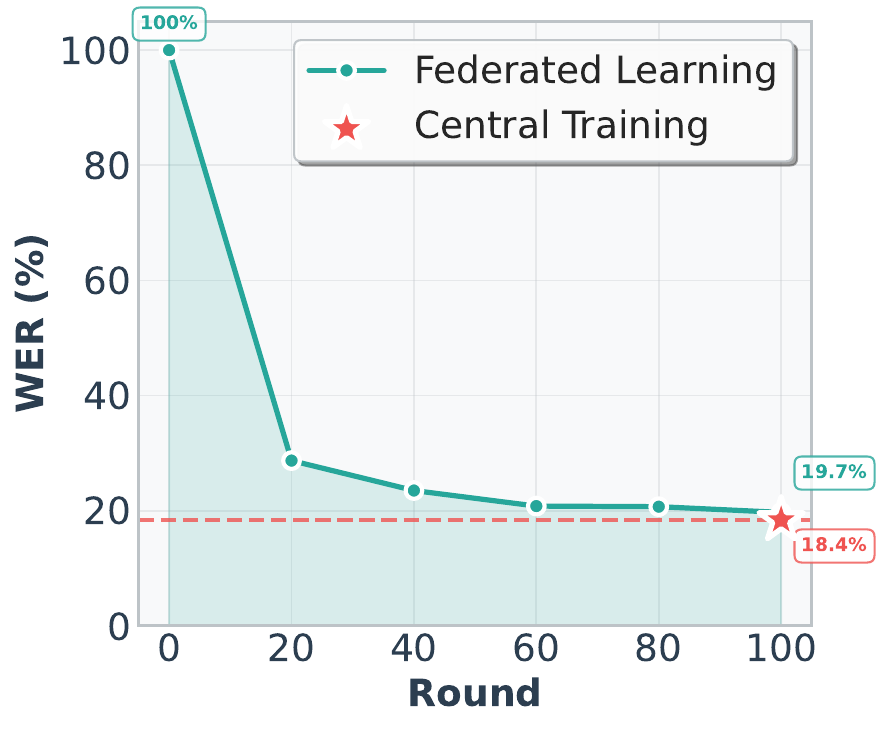} }}%
    \caption{WER  in Multilingual scenario with SpeechLLM in both central and Federated scenarios using WavLM as speech encoder. (a) LS English dataset. (b) MLS Italian dataset.}%
    \label{fig:multi}%
\end{figure*}

This scenario leverages WavLM as the speech encoder using all available speakers from both LS and MLS Italian, with a total of 316 speakers available during training rounds. Figure ~\ref{fig:multi} illustrates the WER performance across communication rounds for both datasets. Starting from an untrained initialization with 100\% WER, both datsets demonstrate rapid error reduction in early rounds followed by gradual convergence. By round~100, the federated model achieves 16.8\% WER on LS (Figure ~\ref{fig:multi}(a)) and 19.7\% WER on MLS Italian (Figure ~\ref{fig:multi}(b)). LibriSpeech exhibits greater variability across rounds with notable fluctuations between rounds~20 and~60, while MLS Italian demonstrates more stable monotonic convergence. 

Compared to centralized training (6.1\% WER on LS and 18.4.\% WER on MLS Italian), the federated approach shows performance gaps of 10.7 percentage points for LibriSpeech and a negligible 0.3 percentage points (actually outperforming) for MLS Italian. 

Overall, these results demonstrate effective federated convergence leveraging all available clients, with the multilingual model achieving near-parity with centralized training on Italian while maintaining a moderate gap on English.

\section{Conclusion and Future Work}
This paper presented a systematic study of federated training for SpeechLLM-based end-to-end ASR in both English and Italian monolingual settings. By integrating parameter-efficient adaptation (LoRA and projector layers) within a federated framework, we demonstrated that large speech–language architectures can be effectively optimized without centralized data aggregation. The proposed Adaptive FedAvg strategy improved convergence stability and reduced the performance gap between centralized and federated training. Experimental results showed that federated SpeechLLM achieves near-centralized performance on LibriSpeech and maintains competitive accuracy on MLS Italian, while significantly reducing communication overhead by updating only a small fraction of model parameters. Furthermore, the comparison between WavLM and Whisper encoders highlights the importance of pretrained multilingual robustness in federated environments. Overall, our findings establish SpeechLLM combined with PEFT as a practical and scalable solution for privacy-preserving ASR.

Future research will extend this framework toward large-scale multilingual and cross-domain federated ASR scenarios involving more diverse acoustic conditions and highly non-IID client distributions. Incorporating advanced communication-efficient techniques such as gradient compression, adaptive client selection, or hierarchical aggregation could further improve scalability. Additionally, integrating differential privacy mechanisms would provide formal privacy guarantees beyond data locality. Exploring personalization strategies, including client-specific adapters or federated continual learning, may help mitigate heterogeneity and close the remaining centralized federated performance gap. Finally, scaling to larger LLM backbones and evaluating real-world on-device deployment constraints will be essential for practical federated SpeechLLM applications.

\bibliographystyle{IEEEtran}
\bibliography{mybib}

\end{document}